\documentclass[journal]{IEEEtran}
\usepackage{times}
\usepackage{epsfig}
\usepackage{graphicx}
\usepackage{amsmath}
\usepackage{amssymb}
\usepackage{booktabs,subfigure,color,threeparttable}
\graphicspath{{./figures/}}
\usepackage{tabularx}
\usepackage{multirow}
\usepackage{tabulary}

\ifCLASSINFOpdf
\else
   \usepackage[dvips]{graphicx}
\fi
\usepackage{url}

\usepackage{graphicx}

\begin{document}

\title{Learning Semantically Enhanced Feature for Fine-Grained Image Classification}

\author{Wei Luo*, \IEEEmembership{Member, IEEE}, Hengmin Zhang*, Jun Li, and Xiu-Shen Wei

\thanks{Submitted date: 07/05/2020. This work was supported in part by NSFC under grant 61702197, in part by NSFGD under grant 2020A151501813 and 2017A030310261.}
\thanks{Wei Luo is with the South China Agricultural University, Guangzhou, 510000, China (email:cswluo@gmail.com).}
\thanks{Hengmin Zhang is with the East China University of Science and Technology, Shanghai, 200237, China (email:zhanghengmin@126.com).}
\thanks{Jun Li and Xiu-Shen Wei are with the Nanjing University of Science and Technology, Nanjing, 210094, China (email:\{junli,weixs\}@njust.edu.cn).}
\thanks{* indicates equal contribution. Wei Luo is the corresponding author.}
}

\markboth{IEEE SIGNAL PROCESSING LETTERS. VOL. 27, 2020}
{Shell \MakeLowercase{\textit{et al.}}: Bare Demo of IEEEtran.cls for IEEE Journals}
\maketitle

\begin{abstract}
We aim to provide a computationally cheap yet effective approach for fine-grained image classification (FGIC) in this letter. 
Unlike previous methods that rely on complex part localization modules,
our approach learns fine-grained features by enhancing the semantics of sub-features of a global feature. 
Specifically, we first achieve the sub-feature semantic by arranging feature channels of a CNN into different groups through channel permutation. 
Meanwhile, to enhance the discriminability of sub-features, the groups are guided to be activated on object parts with strong discriminability by a weighted combination regularization. 
Our approach is parameter parsimonious and can be easily integrated into the backbone model as a plug-and-play module for end-to-end training with only image-level supervision.
Experiments verified the effectiveness of our approach and validated its comparable performance to the state-of-the-art methods. Code is available at {\it \url{https://github.com/cswluo/SEF}}
\end{abstract}


\IEEEpeerreviewmaketitle

\section{Introduction}
\label{sec:intro}
\IEEEPARstart{F}{ine}-grained image classification (FGIC) concerns the task of distinguishing subordinate categories of some base classes such as dogs~\cite{stdogs11feifei}, birds~\cite{cubbirds11caltech}, cars~\cite{stcars13feifei}, aircraft~\cite{vggaircraft13Vedaldi}. Due to the large intra-class pose variation and high inter-class appearance similarity, as well as the scarcity of annotated data, it is challenging to efficiently solve the FGIC problem. 

Recent studies have shown great interests in tackling the FGIC problem by unifying part localization and feature learning in an end-to-end CNN~\cite{racnn@mei,macnn@mei,dfbnet18larry,mamc18eccv,trilinear_attention@luojiebo,s3n@19iccv,dbtnet@19nips}. 
For example,~\cite{fcan@lin,ntscnn@eccv}, and \cite{racnn@mei} first crop object parts on input images and feed them into models for feature extraction, where the part locations are obtained by taking as input the image-level features extracted by a convolutional network. 
To make the model optimization more easier,~\cite{mamc18eccv} and \cite{crossx@luowei} produce parts features by weighting feature channels using soft attentions. 
Another line of research divides feature channels into several groups with each corresponding to a semantic part of the input image~\cite{macnn@mei,trilinear_attention@luojiebo}. 
However, these methods usually make the model optimization more difficult, since they either rely on complex part localization modules, or introduce a large number of parameters into their backbone models, or require a separate module to guide the learning of the feature channel grouping. 


\begin{figure*}[t]
\centering
\includegraphics[width=0.95\linewidth]{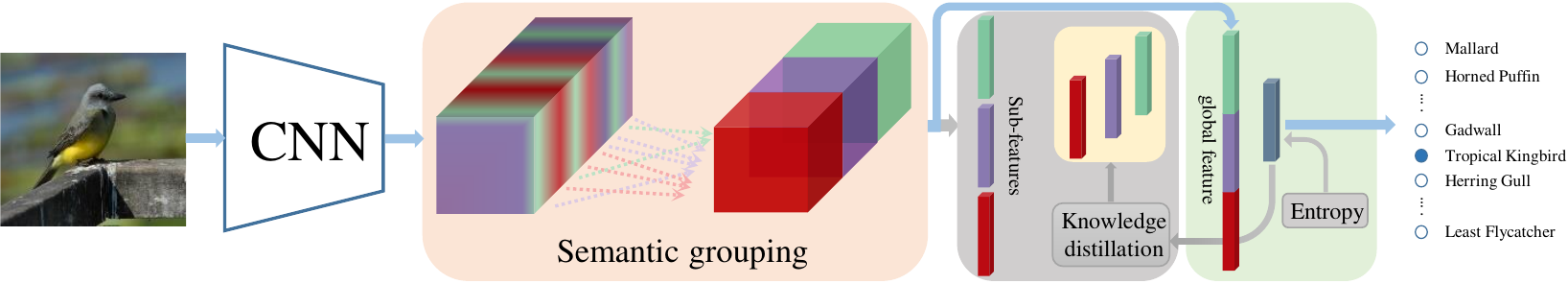}
\caption{Overview of our approach. The last-layer convolutional feature channels (depicted by the mixed color block) of the CNN are arranged into different groups (represented by different colors) by our semantic grouping module. The global and its sub-features (group-wise features) are obtained from the arranged feature channels by average pooling. The light yellow block in the gray block denotes the predicted class distributions from corresponding sub-features, which are regularized by the output of the global feature through knowledge distillation. All gray blocks are only effective in the training stage while removed in the testing stage. The details of the CNN are omitted for clarity. Best viewed in color. }
\label{fig:overview}
\end{figure*}

In this letter, we propose a computationally cheap yet effective approach that learns fine-grained features by improving the sub-features' semantics and discriminability.
It includes two core components: 
{\it 1)} a semantic grouping module that arranges feature channels with similar properties into the same group to represent a semantic part of the input image; 
{\it 2)} a feature enhancement module that improves the discriminability of the grouped features by guiding them to be activated on object parts with strong discriminability. 
The two components can be easily implemented as a plug-and-play module in modern CNNs without the need of guided initialization or modifying the backbone network structure. 
By coupling the two components together, our approach ensures the generation of powerful and distinguishable fine-grained features without requiring complex part localization modules.

Concretely, we construct semantic groups in the last convolutional layer of a CNN by arranging feature channels through a permutation matrix,       
which is learned implicitly through regularizing the relationship between feature channels, i.e., maximizing the correlations between feature channels in the same predefined group and decorrelating those in different predefined groups. Thus, our feature channel grouping does not introduce any additional parameters into the backbone model.
Compared to~\cite{trilinear_attention@luojiebo} and \cite{macnn@mei}, our strategy groups feature channels more consistently and requires no guided initialization.
To guide the semantic groups to be activated on object parts with strong discriminability, we introduce a regularization method that employs a weighted combination of maximum entropy learning and knowledge distillation. 
The strategy of weighted combination is derived from matching prediction distributions between the outputs of the global feature and its group-wise sub-features. 
This regularization method introduces only a small number of parameters into the backbone models due to the output of prediction distributions of sub-features.
Overall, by coupling the two components together, our approach can efficiently obtain fine-grained features with strong discriminability. 
Besides, our approach can be easily integrated into the backbone model as a plug-and-play module for end-to-end training with only image-level supervision. 
Our contributions are summarized as follows:
\begin{itemize}
    \item[$\bullet$] We propose a computationally cheap FGIC approach that achieves comparable performance to the state-of-the-art methods with only $\mathbf{1.7\%}$ parameters more than its ResNet-50 backbone on the Birds dataset.
    \item[$\bullet$] We propose to achieve part localization by learning semantic groups of feature channels, which does not require guided initialization and extra parameters.
    \item[$\bullet$] We propose to enhance feature discriminability by guiding its sub-features to be extracted from object parts with strong discriminability.
\end{itemize}

\section{Proposed Approach}
\label{sec:model}
Our approach involves two main components: 1) A semantic grouping module that arranges feature channels with different properties into different groups. 
2) A feature enhancement module that elevates feature performance by improving its sub-features discriminability. Fig.~\ref{fig:overview} is an overview of our approach.

\subsection{Semantic Grouping}
\label{sec:group}
Previous work \cite{net2vec@18cvpr} has verified that bunches of filters in the high-level layers of CNNs are required to represent a semantic concept. Therefore, we develop a regularization method that arranges filters with different properties into different groups to capture semantic concepts. Specifically, given a convolutional feature $\mathbf{X}^L\in \mathbb{R}^{C\times WH}$ of layer $L$, where a single feature channel is represented by $\mathbf{X}^L_i\in \mathbb{R}^{WH}, i\in [1,\cdots,C]$. 
We first arrange its feature channels through a permutation operation $\mathbf{X}^{L'}=\mathbf{AX}^L$, where $\mathbf{A} \in \mathbb{R}^{C\times C}$ is a permutation matrix, and then divide the channels into $G$ groups. Since $\mathbf{X}^L$ is obtained by convolving the filters of layer $L$ with the features of layer $L-1$. The convolution operation can be formulated as
\begin{equation}
\label{eq:convle_to_mm}
    \mathbf{X}^L=\mathbf{BX}^{L-1},
\end{equation}
where $\mathbf{B}\in \mathbb{R}^{C\times \Omega}$ and $\mathbf{X}^{L-1}\in \mathbb{R}^{\Omega\times \Psi}$ are respectively the reshaped filters of layer $L$ and the reshaped feature of layer $L-1$. Thus $\mathbf{X}^{L'}$ can be rewritten as
\begin{equation}
\label{eq:mm_to_convolve}
    \mathbf{X}^{L'}=\mathbf{AX}^L=\mathbf{ABX}^{L-1}=\mathbf{WX}^{L-1},
\end{equation}
where $\mathbf{W}$ is a permutation of $\mathbf{B}$.

To achieve the groups with semantic meaning, $\mathbf{A}$ should be learned to discover the similarities between the filters (rows) of $\mathbf{B}$. It is, however, nontrivial to learn the permutation matrix straightforwardly. Therefore, we instead learn $\mathbf{W}$ directly by constraining the relationships between feature channels of $\mathbf{X}^{L'}$, thus circumventing the difficulty of learning $\mathbf{A}$. 
To effect, {\it we maximize the correlation between feature channels in the same group while decorrelating those in different groups}. Concretely, let $\tilde{\mathbf{X}}^{L'}_{i} \leftarrow \mathbf{X}^{L'}_{i} / ||\mathbf{X}^{L'}_{i}||_2 $ be a normalized channel. The correlation between channels is then defined as
\begin{equation}
\label{eq:correlation}
    d_{ij} = {\tilde{\mathbf{X}}_{i}^{L'^T}} \tilde{\mathbf{X}}^{L'}_{j},
\end{equation}
where $T$ is the transpose operation. Let $\mathbf{D}\in \mathbb{R}^{G\times G}$ be the correlation matrix with element $\mathbf{D}_{mn}=\frac{1}{C_mCn}\sum_{i\in m, j\in n}d_{ij}$ corresponding to the average correlation of feature channels from groups $m$ and $n$, where $m, n \in {1,\cdots, G}$ and $C_m$ is the number of channels in group $m$. Then the semantic groups can be achieved by minimizing
\begin{equation}
\label{eq:group_loss}
    \mathcal{L}_{group} = \frac{1}{2}(\lVert \mathbf{D} \rVert_F^2 - 2 \lVert diag(\mathbf{D}) \rVert_2^2),
\end{equation}
where $diag(\cdot)$ extracts the main diagonal of a matrix.

Generally, Eq.~\ref{eq:mm_to_convolve} can be implemented as a convolutional operation in CNNs. Eq.~\ref{eq:group_loss} can be regularized on feature channels without introducing any additional parameters.
Different from~\cite{dbtnet@19nips} in which a semantic mapping matrix is explicitly learned to reduce the redundancy of bilinear features, our strategy focuses on improving the semantics of sub-features and does not modify the backbone network structure. 

\subsection{Feature Enhancement}
\label{sec:enhance}
Semantic grouping can drive features of different groups to be activated on different semantic (object) parts. However, the discriminability of those parts may not be guaranteed. We, therefore, need to guide these semantic groups to be activated on object parts with strong discriminability. A simple way to achieve this effect is to match the prediction distributions between the object and its parts as implemented in~\cite{crossx@luowei}. However, it is unclear about the reasons why matching distributions can improve performance. Here, we provide an analysis to understand its principles and make improvements to achieve better performance.

Let $\mathbf{P}_w$ and $\mathbf{P}_a$ be the prediction distributions of an object and its part, respectively. Then, matching their distributions can be achieved by minimizing the KL divergence between them~\cite{crossx@luowei},
\begin{equation}
\label{eq:kl}
    \mathcal{L}_{\text{KL}(\mathbf{P}_w||\mathbf{P}_a)} = - \text{H}(\mathbf{P}_w) + \text{H}(\mathbf{P}_w,\mathbf{P}_a),
\end{equation}
where $\text{H}(\mathbf{P}_w) = -\sum \mathbf{P}_w \log \mathbf{P}_w$ 
and $\text{H}(\mathbf{P}_w,\mathbf{P}_a) = - \sum \mathbf{P}_w \log{\mathbf{P}_a}$. Thus, the optimization objective of a classification task can be generally written as
\begin{equation}
\label{eq:objective}
\begin{split}
    \mathcal{L} = \mathcal{L}_{cr} +\lambda \mathcal{L}_{KL},
\end{split}
\end{equation}
where $\mathcal{L}_{cr}$ is the cross entropy loss and $\lambda$ is the balance weight. Substituting Eq.~\ref{eq:kl} into Eq.~\ref{eq:objective} we have
\begin{equation}
\label{eq:final_objective}
    \mathcal{L} = \mathcal{L}_{cr} - \lambda \text{H}(\mathbf{P}_w) + \lambda \text{H}(\mathbf{P}_w,\mathbf{P}_a).
\end{equation}
Eq.~\ref{eq:final_objective} implies that minimizing matching prediction distributions can be decomposed into a maximum entropy term and a knowledge distillation term. Both of them are powerful regularization methods~\cite{maximumentropy@cl,distill15hinton}. In FGIC, maximum entropy learning can effectively reduce the confidence of classifiers, thus leading to better generalization in low data-diversity scenarios~\cite{maxent@18nips}. 
The last term in Eq.~\ref{eq:final_objective} distills knowledge from the global feature to the local feature,  thus enhancing the discriminability of local features. To this end, we can regulate the importance of the two terms separately for better performance. Put all together, our optimization objective can be formulated as
\begin{equation}
    \mathcal{L}=\mathbb{E}_{\mathbf{x}} \bigl(\mathcal{L}_{cr} - \lambda \text{H}(\mathbf{P}_w) + \frac{\gamma}{G} \sum_{g=1}^G \text{H}(\mathbf{P}_w,\mathbf{P}_a^g) + \phi \mathcal{L}_{group}\bigr).
\end{equation}
Here, $\lambda, \gamma, \phi$ and $G$ are hyper-parameters and we omit the dependence on $\mathbf{x}$ for clarity (see Table~\ref{tab:ablation} for evaluation).

In our implementation, the last-layer feature channels are first pooled averagely and then simultaneously fed into $G+1$ neural networks (NNs) to output class distributions, in which one takes as input the global feature and the others take as input only features in corresponding groups. Only the NN taking the global feature is used for prediction (see Fig.~\ref{fig:overview}). 

\section{Experiments}
\label{sec:exp}

\subsection{Experimental Setup}
\label{sec:setup}
We employ ResNet-50~\cite{resnet16kaiming} as the backbone of our approach in PyTroch and experiment on CUB-Birds~\cite{cubbirds11caltech}, Stanford Cars~\cite{stcars13feifei}, Stanford Dogs~\cite{stdogs11feifei}, and FGVC-Aircraft~\cite{vggaircraft13Vedaldi} datasets (see supplementary materials (SMs) for data statistics). We initialize our model using the pretrained weights on ImageNet~\cite{imagenet@feifei} and fine-tune all layers on batches of 32 images of size $448\times 448$ by SGD~\cite{lecun98cnn} with momentum of 0.9. Random flip is employed only in training. The $G+1$ NNs are all single-layer networks. The initial learning rate, $lr$, is $0.01$ except on Dogs where $0.001$ is used and decays by $0.1$ every $20$ epochs with a total of $50$ training epochs. $\lambda, \gamma$ and $\phi$ are validated on the Birds validation set, which contains $10\%$ of the training samples, and are correspondingly set to $1, 0.05$ and $1$ across all datasets. $G$ is determined by the performance of models with different values and respectively set to $3, 4$, and $2$ on Aircraft, Birds, and the other two datasets in this letter.

\textbf{Metrics}. Except for classification accuracy, we also employ scores to rank the overall performance of a method across datasets. Given the performance of method $m$ on $L$ datasets, the score of method $m$ is $S_m = \frac{1}{L} \sum_{l=1}^L R_l^m$, where $R_l^m$ is the rank of method $m$ on the $l$-th dataset based on its classification accuracy. The closer the score is to 1, the better.


\subsection{Comparison with the State-of-the-Art}
\label{sec:comparison}
To be fair, we only compare to weakly-supervised methods employing the ResNet-50 backbone, due to its popularity and state-of-the-art performance in recent FGIC work. {\it Notice that we are not attending to achieve the best performance but to emphasize the advantage brought by our simple construction}.

\textbf{Complexity analysis.} Our approach introduces additional parameters into the backbone network only in knowledge distillation, where the total number of additional parameters is constrained by the multiplication of feature dimensionality and the number of classes. For example, $409,600$ new parameters are introduced on the Birds dataset, which accounts for only $\mathbf{1.7\%}$ of the parameters of ResNet-50. Since there are only negligible additional parameters in our approach, the network is efficient to train. Compared with computation-intensive methods such as S3N~\cite{s3n@19iccv}, TASN~\cite{trilinear_attention@luojiebo}, API-Net~\cite{attentive_pairwise_interaction@20aaai}, and DCL~\cite{dcl@meitao} (requires 60, 90, 100, and 180 epochs for training, respectively), our approach can be optimized in 50 epochs.

During testing, only the backbone network is activated with all additional modules removed. Compared with its ResNet-50 backbone, our approach boosts performance $\mathbf{+1.65\%}$ on average with the same time cost at inference, which indicates the practical value of our approach.
\begin{table}[h]
    \centering
    \caption{Comparison with state-of-the-art methods ($\%$).}
    \begin{threeparttable}
    \begin{tabular}[width=\linewidth]{lcccc|c}
        \toprule
            & Birds & Cars & Dogs &Aircraft & Scores\\
        \midrule
        Kernel-Pooling$^{\star}$~\cite{kp@17cvpr}           &$84.7$             &$92.4$             &$-$                                &$86.9$             &$10.3$\\
        MAMC-CNN~\cite{mamc18eccv}                          &$86.2$             &$93.0$             &$84.8$                             &$-$                &$7.7$\\
        DFB-CNN$^{\star}$~\cite{dfbnet18larry}              &$87.4$             &$93.8$             &$-$                                &$92.0$             &$6.3$\\
        NTS-Net$^{\dagger}$~\cite{ntscnn@eccv}              &$87.5$             &$93.9$             &$-$                                &$91.4$             &$6.0$\\
        
        S3N$^{\dagger}$~\cite{s3n@19iccv}                   &$\mathbf{88.5}$    &$94.7$             &$-$                                &$92.8$             &$1.7$\\
        API-Net~\cite{attentive_pairwise_interaction@20aaai}&$87.7$             &$\mathbf{94.8}$    &$88.3$                             &$\mathbf{93.0}$    &$2.3$\\
        DCL~\cite{dcl@meitao}                               &$87.8$             &$94.5$             &$-$                                &$\mathbf{93.0}$    &$2.7$\\
        TASN$^{\dagger}$~\cite{trilinear_attention@luojiebo}&$87.9$             &$93.8$             &$-$                                &$-$                &$5.0$\\
        Cross-X~\cite{crossx@luowei}                        &$87.7$             &$94.6$             &$\mathbf{88.9}$                    &$92.6$             &$2.8$\\
        \midrule
        ResNet-50$^{\ddagger}$~\cite{resnet16kaiming}       &$84.5$             &$92.9$             &$88.1$                             &$90.3$             &$8.3$\\
        MaxEnt-CNN$^{\ddagger}$~\cite{maxent@18nips}        &$83.4$             &$92.8$             &$88.0$                             &$90.7$             &$8.8$\\
        DBT-Net~\cite{dbtnet@19nips}                        &$87.5$             &$94.1$             &$-$                                &$91.2$             &$5.7$\\
        \midrule
        SEF (ours)                                          &$87.3$             &$94.0$             &$88.8$                             &$92.1$             &$4.8$\\
        \bottomrule
    \end{tabular}
    \begin{tablenotes}[flushleft]
        \item {\small $^{\star}, ^{\dagger}$ and $^{\ddagger}$ represent methods with separated initialization, multi-cropping operations, and results from our re-implementation, respectively. The closer the score is to 1, the better.}
    \end{tablenotes}
    \end{threeparttable}
    \label{tab:comparison}
\end{table}

\textbf{Performance comparison.} Table~\ref{tab:comparison} shows that there is no single method that can achieve the best performance on all datasets. Our approach (SEF) attains a score of $\mathbf{4.8}$ ranking $\mathbf{5}$th in overall performance, which is comparable to the state-of-the-art methods, especially considering its simple construction. The methods in the second group are closely related to ours. Compared to ResNet-50 and MaxEnt-CNN, SEF boosts performance on all datasets. 
Besides, SEF achieves a more robust performance than DBT-Net. 
Among other methods, S3N, API-Net, DCL, and Cross-X rank before ours. However, they are more expensive than ours such as Cross-X using twice as many parameters as ResNet-50.

\begin{table}[h]
    \centering
    \caption{Performance on different backbones ($\%$).}
    \begin{threeparttable}
    \begin{tabular}[width=\linewidth]{lcccc|c}
        \toprule
            & VGG16 (+SEF) & ResNet-50 (+SEF) & ResNeXt-50 (+SEF) \\
        \midrule
        Birds   &77.7 (81.1)      &84.5 (87.3)          &86.8 (87.8) \\
        Cars    &87.3 (88.3)      &92.9 (94.0)          &93.7 (94.2)\\
        Dogs    &71.1 (75.4)      &88.1 (88.8)          &89.8 (90.8)\\
        Aircraft &87.0 (88.5)     &90.3 (92.1)          &92.0 (92.6)\\
        \bottomrule
    \end{tabular}
    \end{threeparttable}
    \label{tab:backbones}
\end{table}

Table~\ref{tab:backbones} shows the performance of our approach on different backbones. It is worth noting that {\it 1)} we did not cross-validate the hyper-parameters for these new backbones, but to use those validated for ResNet-50, and {\it 2)} the hyper-parameters are determined through cross-validation on Birds and applied to all datasets without modification. Thus better results can be expected than here. However, the performance demonstrates that our approach can be robustly generalized to different backbones and datasets (Please refer to SMs for more details).


\subsection{Ablation Studies}
\label{sec:ablation}
For simplicity, the rest experiments and analyses are performed on ResNet-18 unless otherwise clarified.

\textbf{Effectiveness of individual module.}
Table~\ref{tab:ablation} shows the results of our approach with different configurations. It reveals that learning semantic groups (row 2) or matching distributions (row 3) independently can slightly improve the performance. Combining both directly (row 4) brings some advantages, 
but the improvement is not systematically consistent. The performance, however, can be significantly improved by decomposing matching distributions into separate regularizers (row 5), which indicates the effectiveness of feature enhancement that guiding semantic groups to be activated on object parts with strong discriminability. 
\begin{figure}[t]
\centering
\includegraphics[width=0.28\columnwidth]{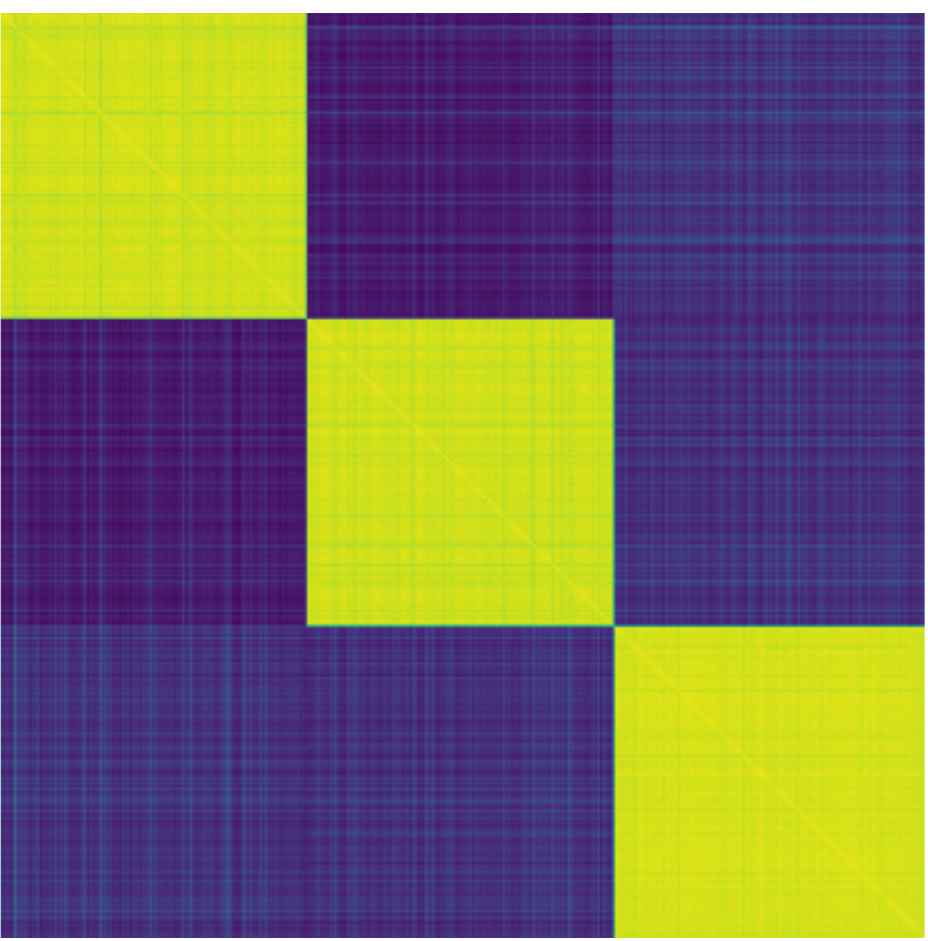} \includegraphics[width=0.28\columnwidth]{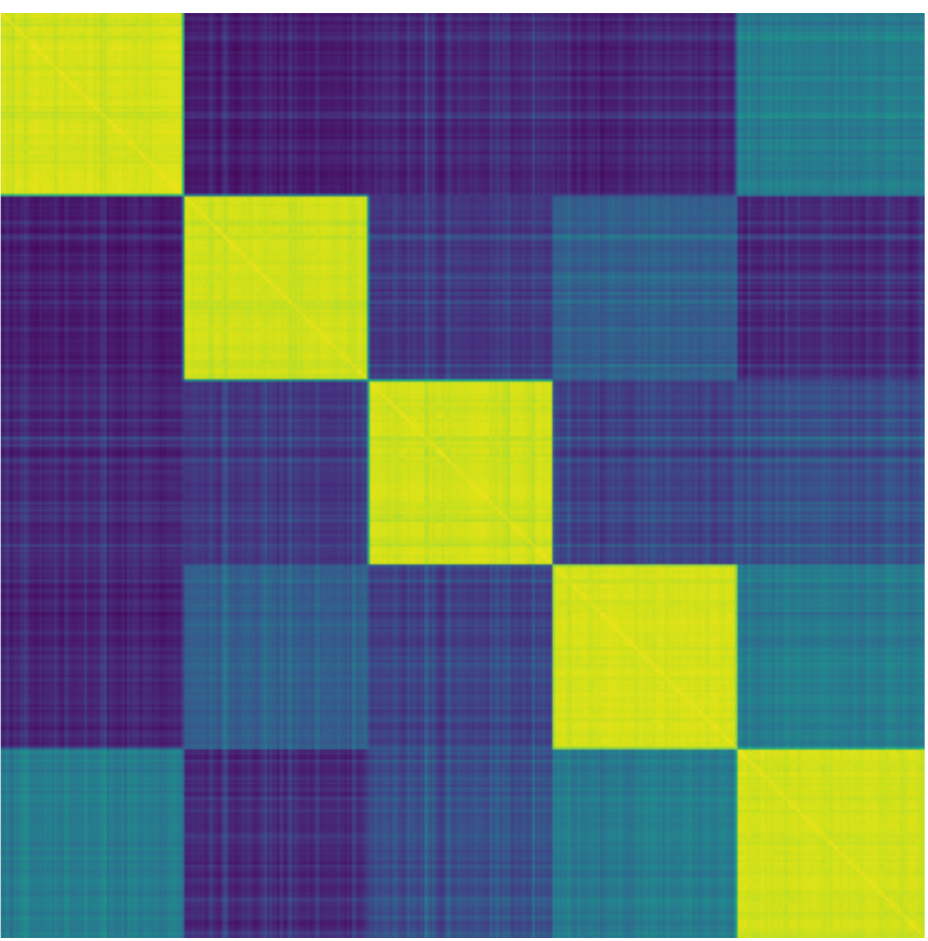} \includegraphics[width=0.335\columnwidth]{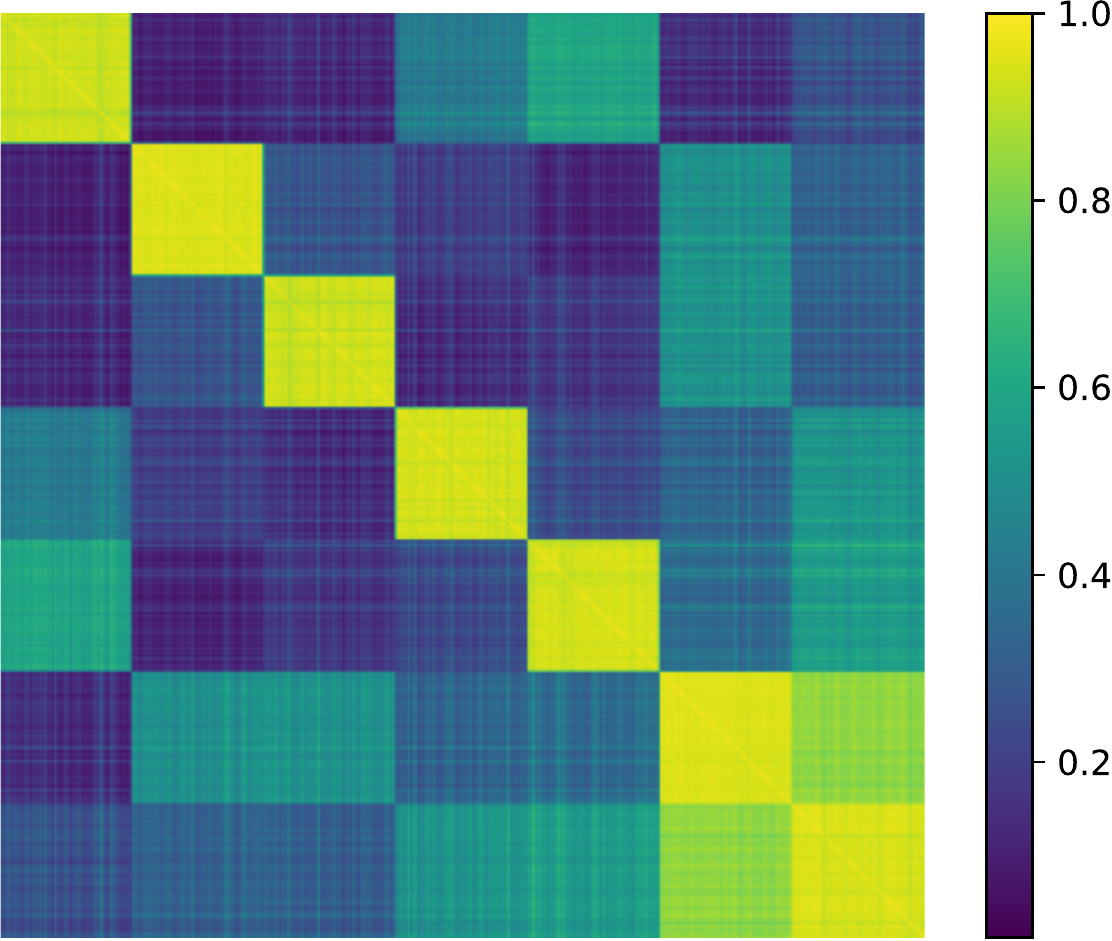} 
\caption{From left to right are correlation matrices of feature channels of models with 3, 5, and 7 groups, averaged on 64 images randomly selected from the Birds testing dataset, respectively.}
\label{fig:correlations}
\end{figure}

\begin{figure}[t]
\centering
\includegraphics[width=0.65\columnwidth]{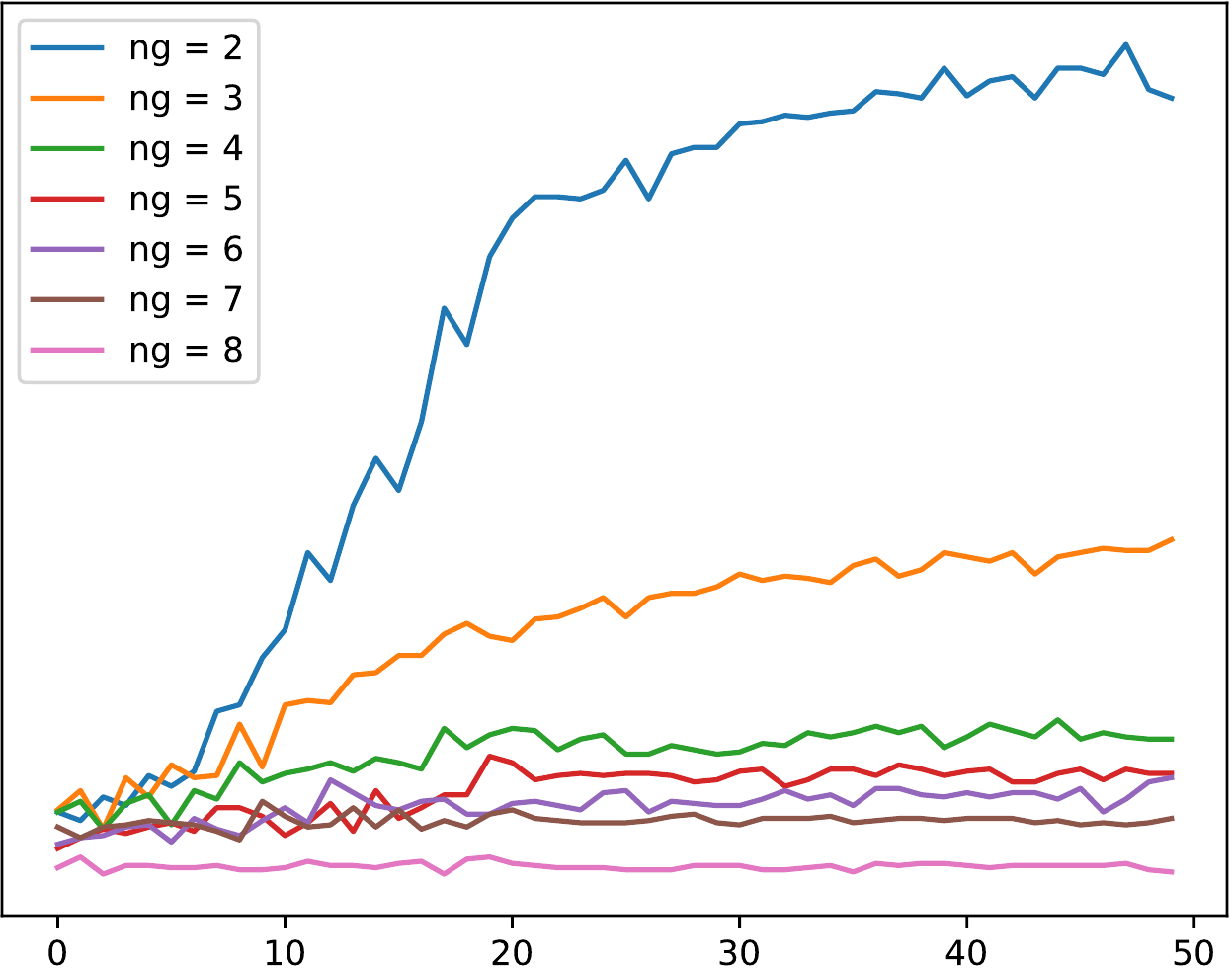}
\caption{Discriminability of the $1$st group features of models at different learning epochs tested on the Birds validation set. ng means the number of groups used in the model.}
\label{fig:ngroups}
\end{figure}

\begin{table}[h]
    \centering
    \caption{Performance with different options ($\%$).}
    \begin{threeparttable}
        \begin{tabular}{lcccc}
        \toprule
                    &Birds & Cars & Dogs &Aircraft \\
        \midrule
        ResNet-18~\cite{resnet16kaiming}            &$82.3$         &$90.4$           &$81.5$           &$88.5$\\
        $\lambda=0, \gamma=0, \phi=1$               &$82.7$         &$90.9$           &$82.1$           &$88.9$\\
        $\lambda=0.05, \gamma=0.05, \phi=0$         &$82.2$         &$90.3$           &$81.8$           &$88.8$\\
        $\lambda=0.05, \gamma=0.05, \phi=1$         &$83.2$         &$90.1$           &$81.8$           &$89.0$\\
        $\lambda=1, \gamma=0.05, \phi=1$            &$\mathbf{84.8}$         &$\mathbf{91.8}$           &$\mathbf{83.1}$           &$\mathbf{89.3}$\\
        \bottomrule
        \end{tabular}
    \end{threeparttable}
    \label{tab:ablation}
\end{table}

\begin{figure}[t]
\begin{center}
\begin{minipage}{0.20\linewidth}
\includegraphics[width=\linewidth]{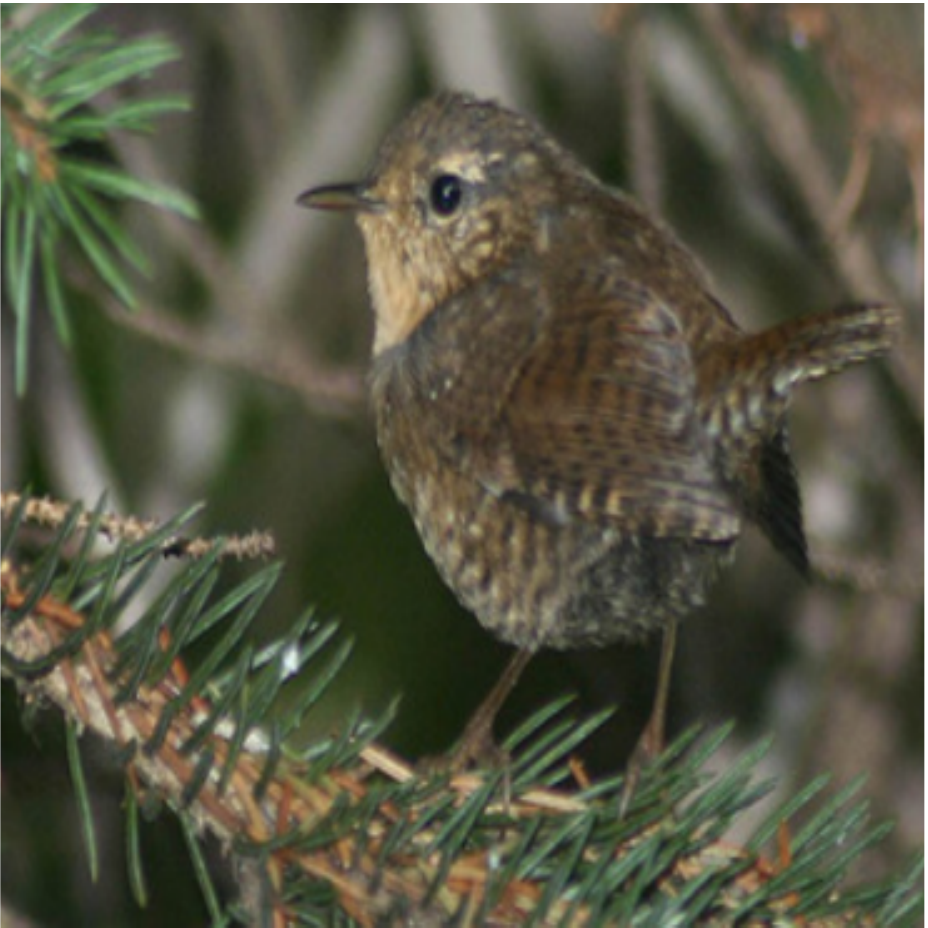}\\
\includegraphics[width=\linewidth]{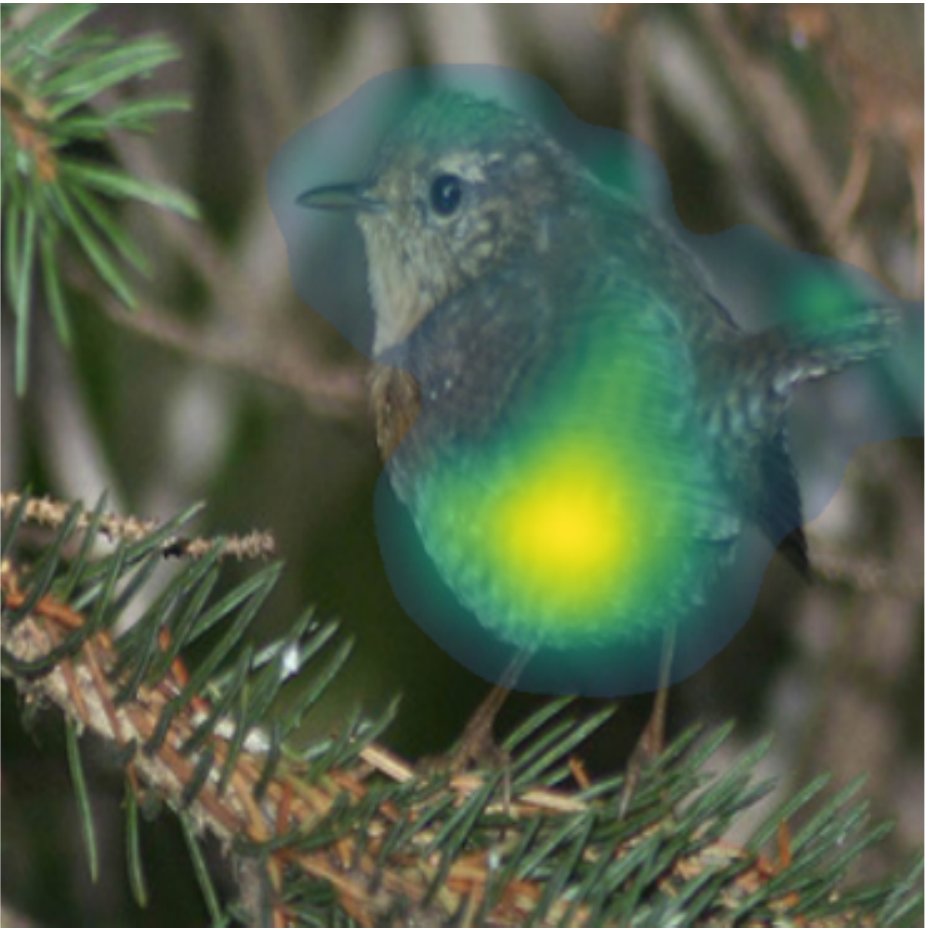}\\
\includegraphics[width=\linewidth]{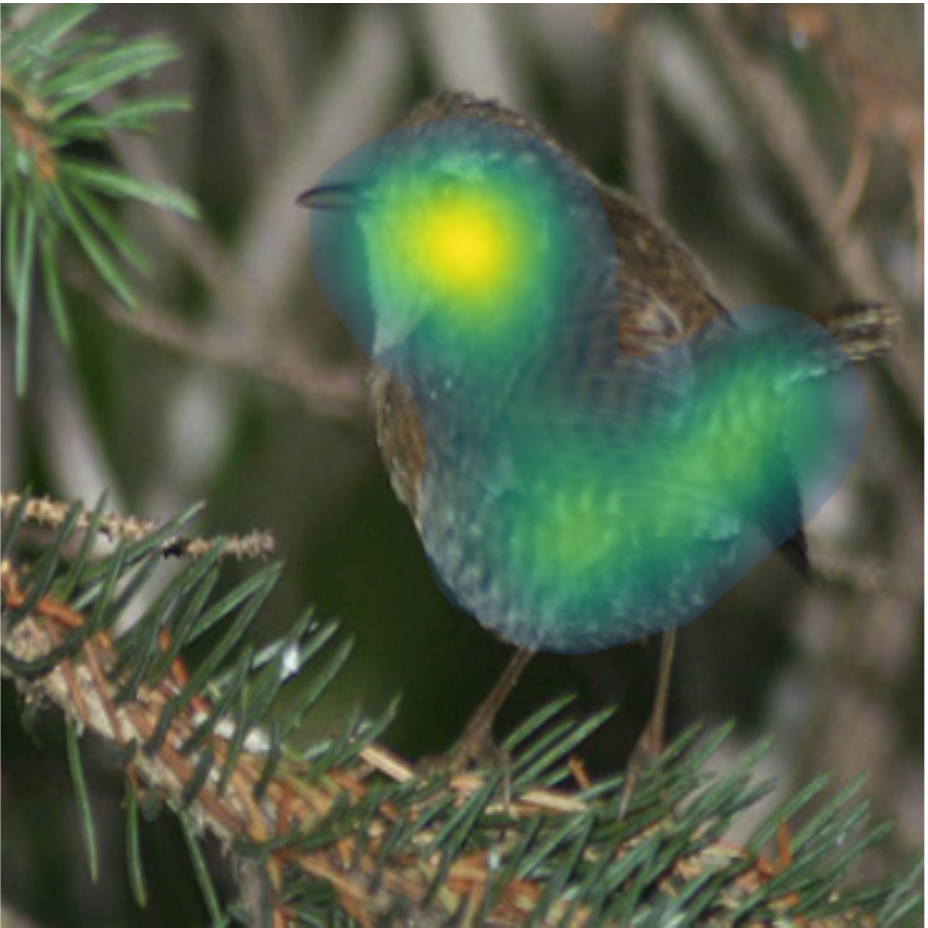}
\end{minipage}
\begin{minipage}{0.20\linewidth}
\includegraphics[width=\linewidth]{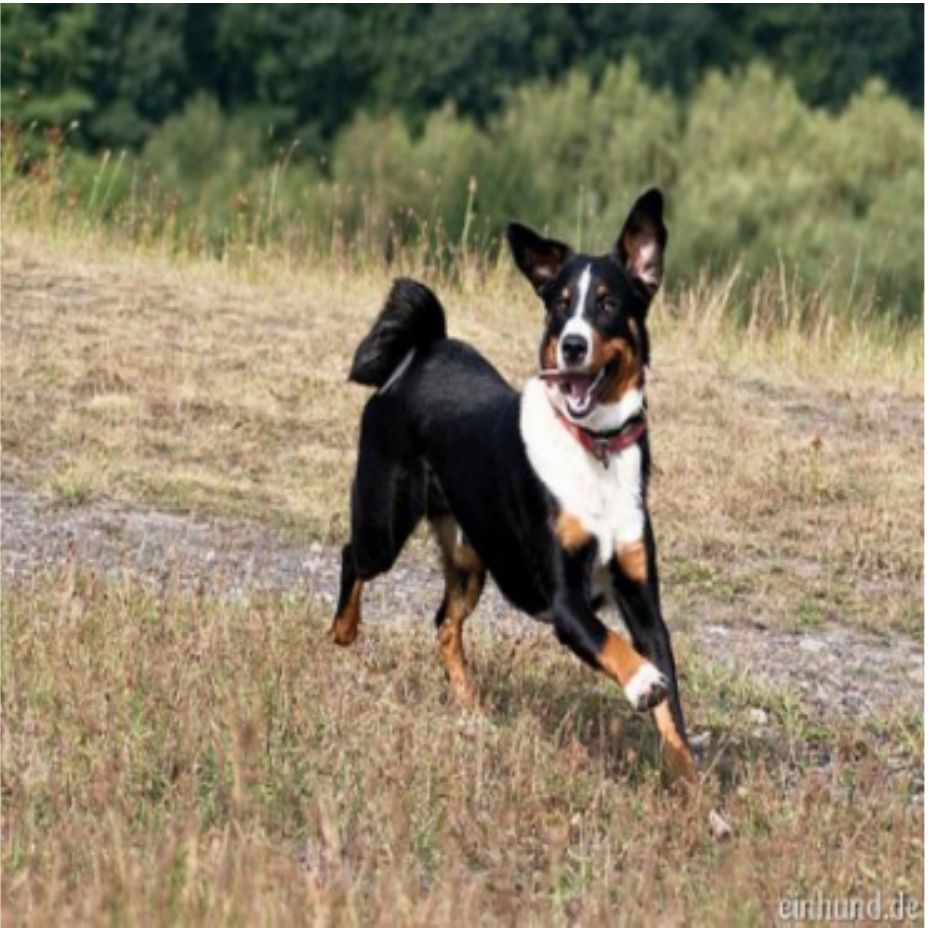}\\
\includegraphics[width=\linewidth]{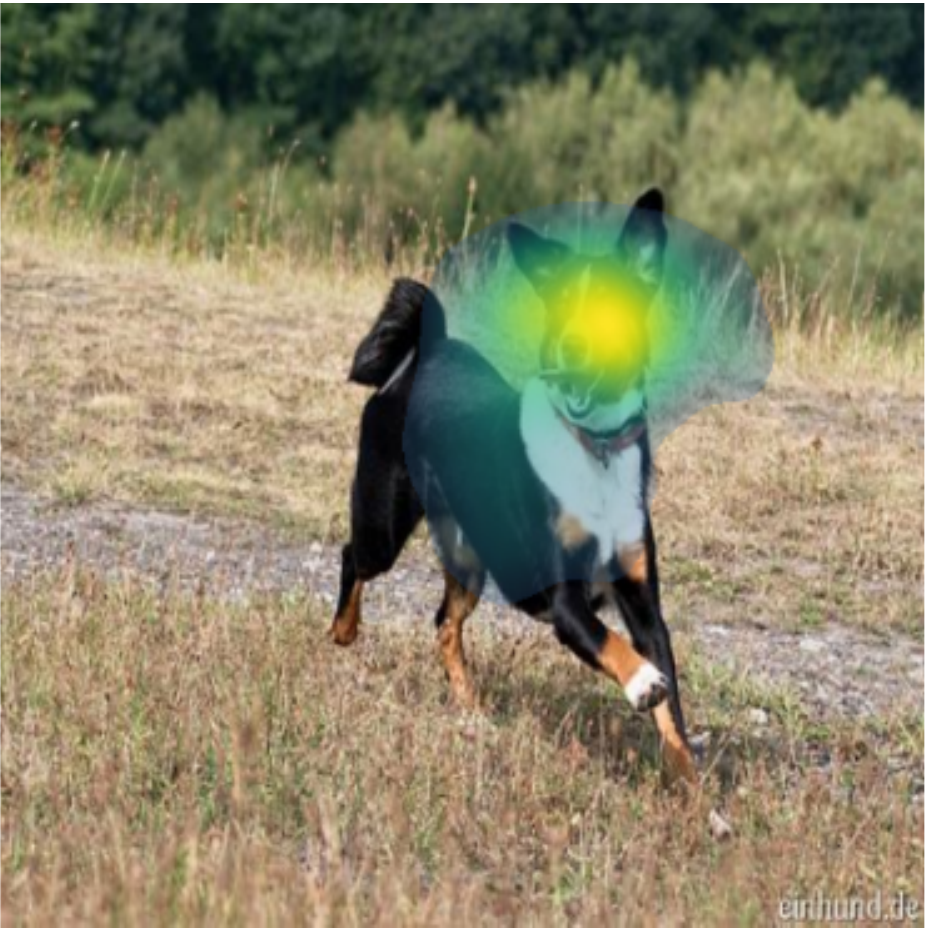}\\
\includegraphics[width=\linewidth]{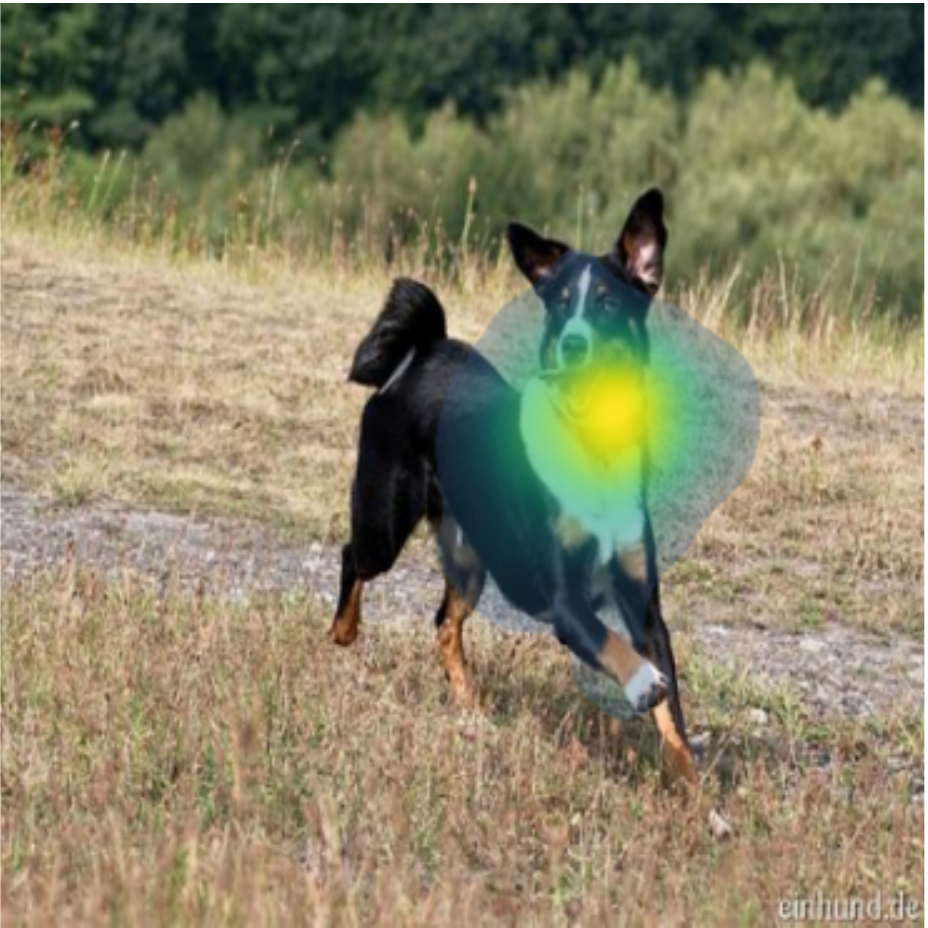}
\end{minipage}
\begin{minipage}{0.20\linewidth}
\includegraphics[width=\linewidth]{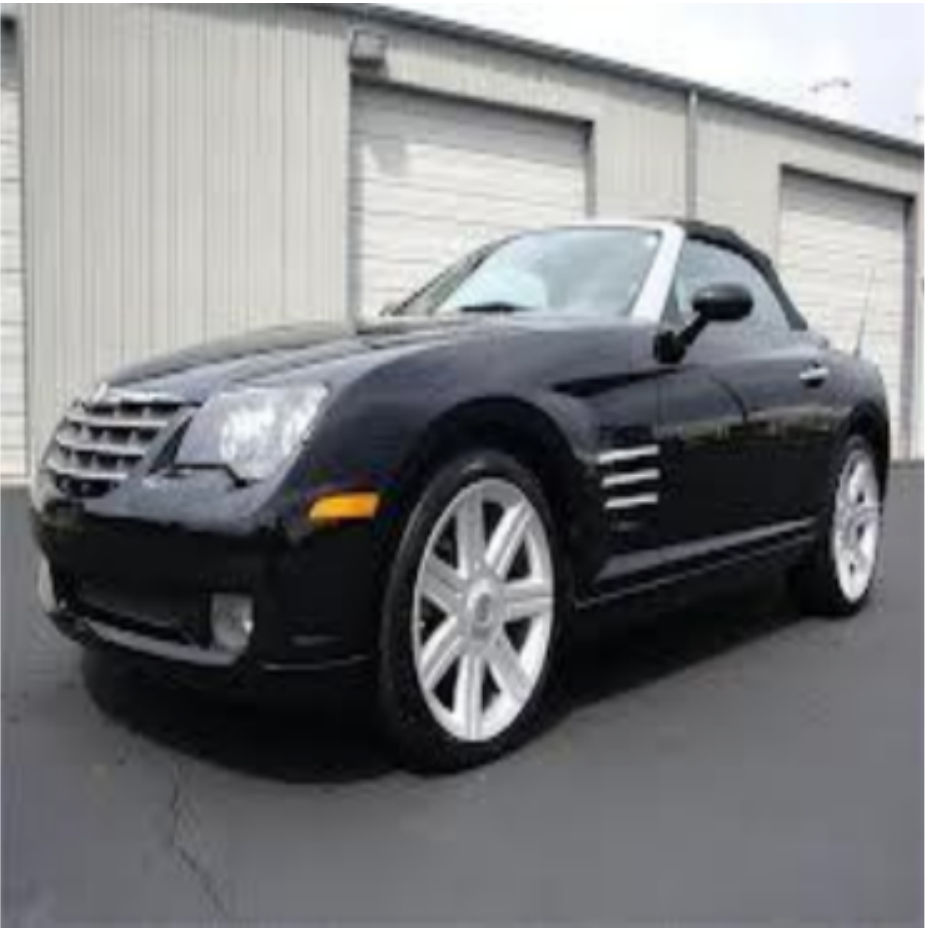}\\
\includegraphics[width=\linewidth]{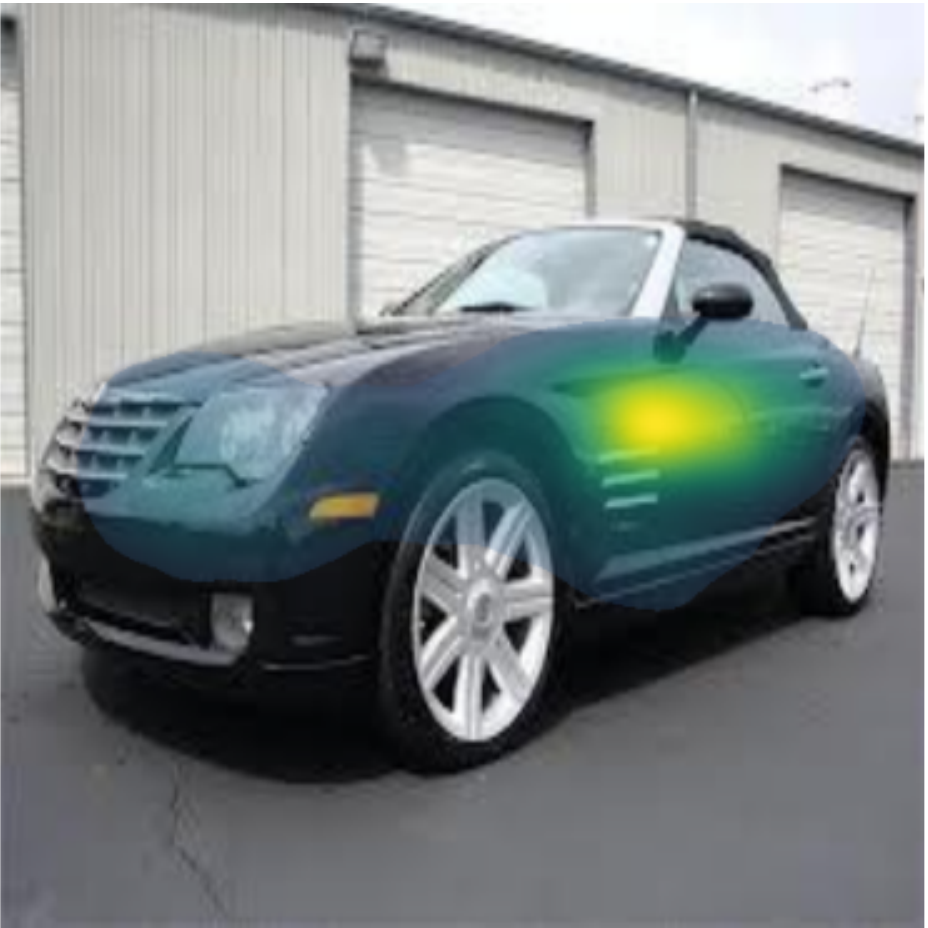}\\
\includegraphics[width=\linewidth]{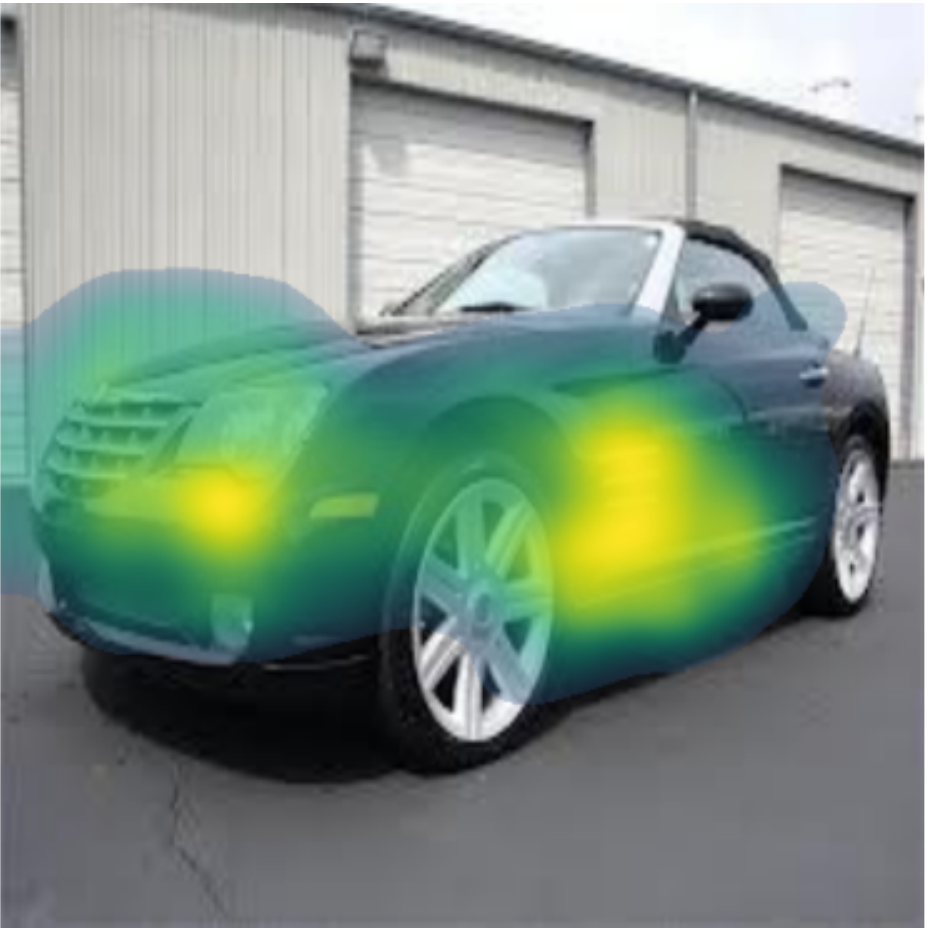}
\end{minipage}
\begin{minipage}{0.20\linewidth}
\includegraphics[width=\linewidth]{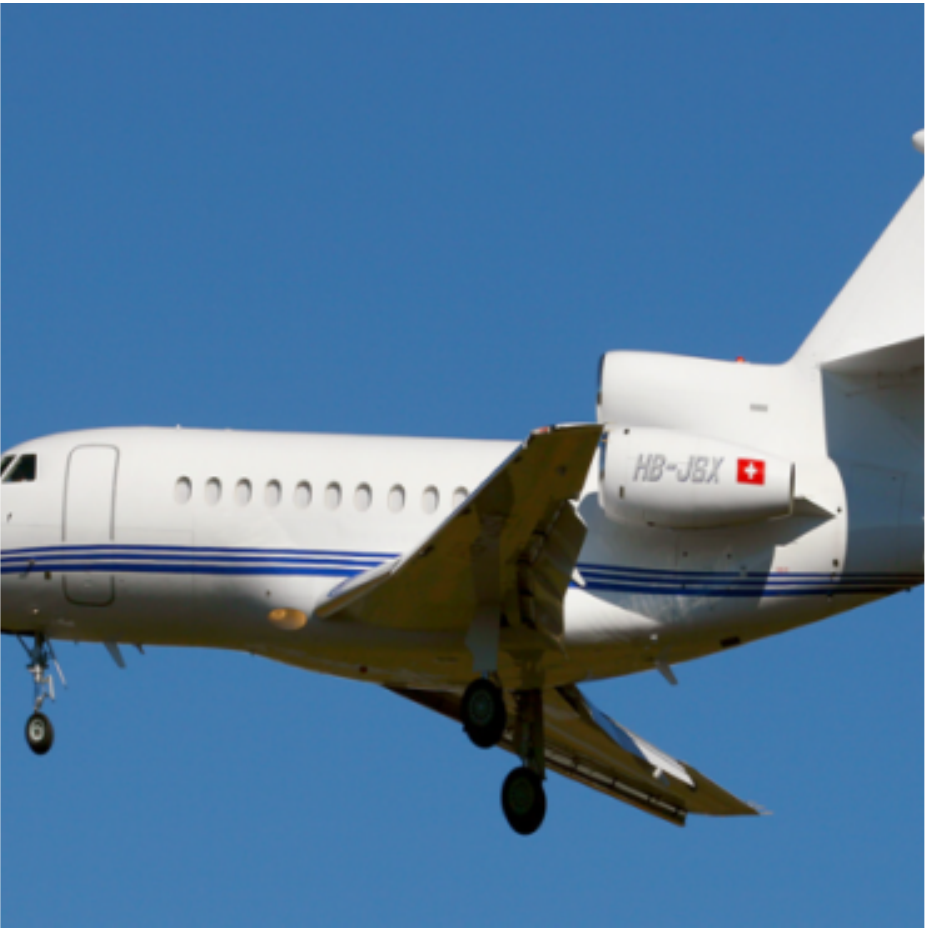}\\
\includegraphics[width=\linewidth]{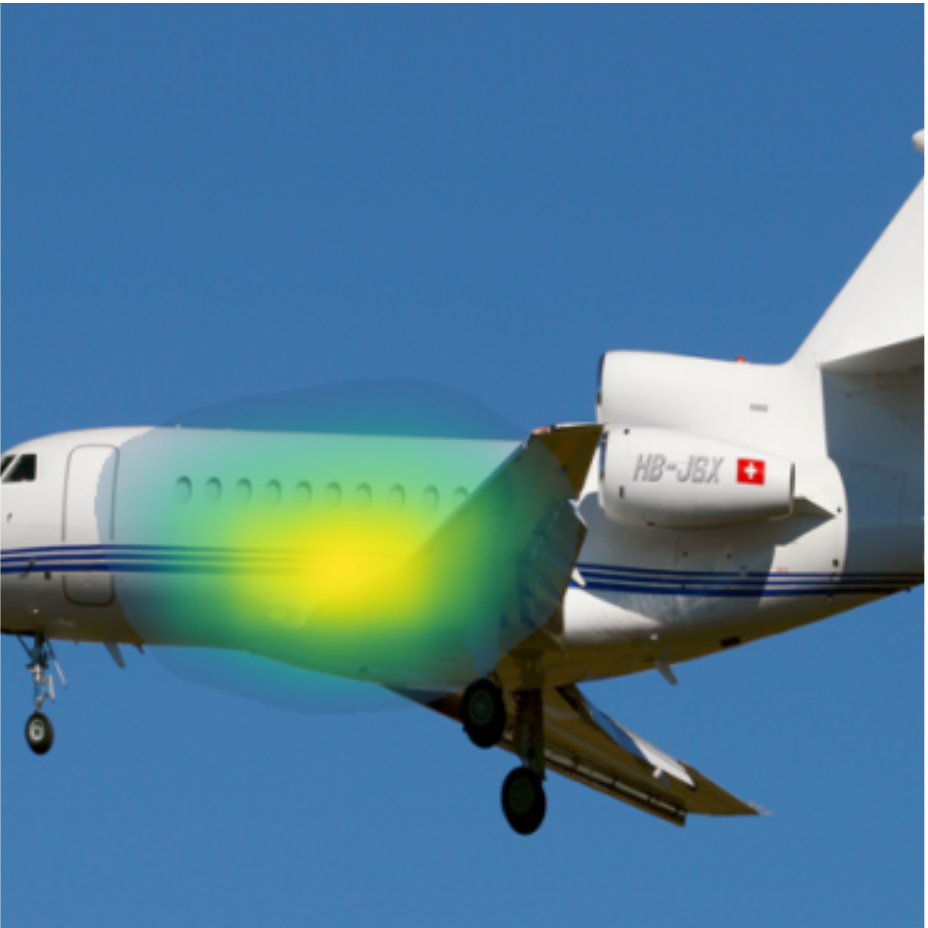}\\
\includegraphics[width=\linewidth]{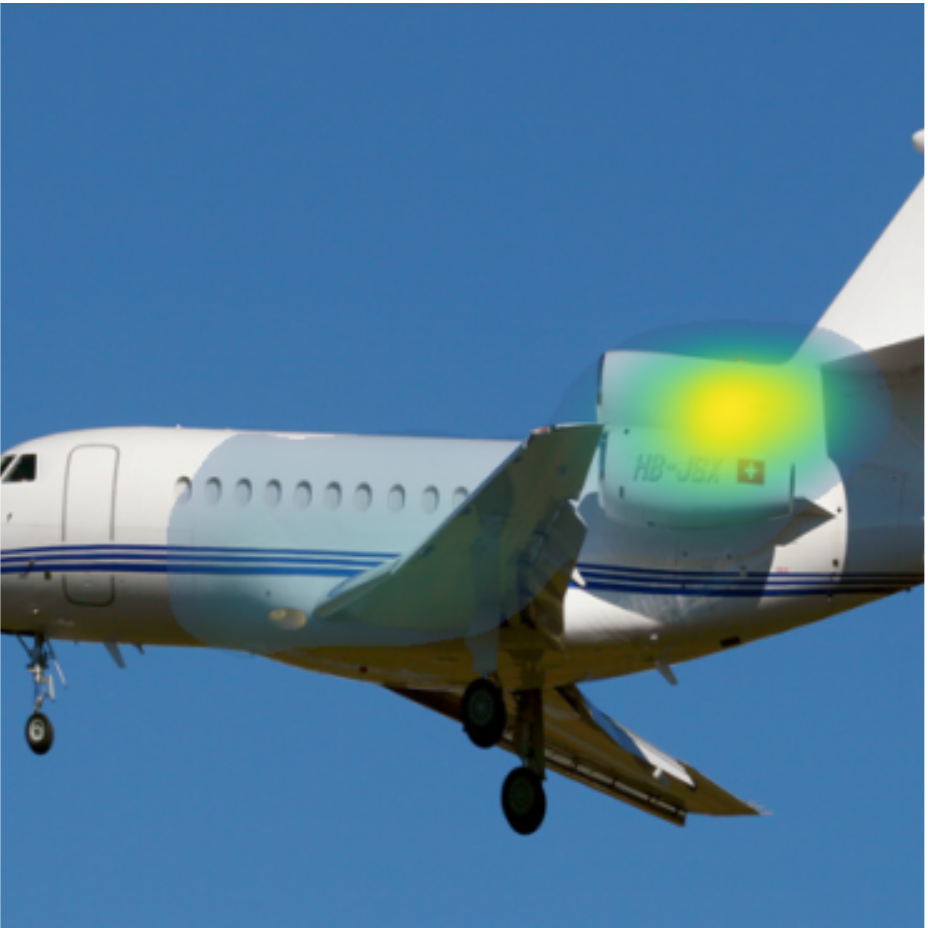}
\end{minipage}
\end{center}
\caption{Group-wise activation maps of models with 2 semantic groups superimposed on original images. The 1st, 2nd, and 3rd rows are respectively the original images, activation maps of the 1st and 2nd semantic groups.}
\label{fig:visualization}
\end{figure}

\textbf{Number of semantic groups.} The group semantic is strongly correlated with its discriminability. However, too many groups may break the correlation, resulting in weak group features. Fig.~\ref{fig:correlations} shows the correlations between feature channels of the last convolutional layer. Fig.~\ref{fig:ngroups} depicts the discriminability of group features of models with varying groups. They illustrate that {\it 1)} our semantic grouping module can effectively achieve the function of grouping correlated feature channels and separating uncorrelated feature channels (Fig.~\ref{fig:correlations}); {\it 2)} feature channels should not be divided into too many groups, which on the one hand causes the difficulty in optimization (Fig.~\ref{fig:correlations}) and, on the other hand, reduces the semantic as well as the discriminability of each group (Fig.~\ref{fig:ngroups}).

\subsection{Visualization}
\label{sec:visualization}
The discriminability of semantic groups (sub-features) can be visualized by highlighting corresponding areas on input images. Ideally, every group should be activated by a set of proximity pixels due to the similar property of neurons in the group. Fig.~\ref{fig:visualization} shows that different groups can be activated by different semantic parts of images, which are highly identifiable such as the wing and engine nacelles of the aircraft. It signifies the success of our approach to enhance feature semantically.


\section{Conclusion}
\label{sec:summary}
In this letter, we proposed a computationally cheap yet effective approach that involves a semantic grouping and a feature enhancement module for FGIC. We empirically studied the effectiveness of each individual module and their combining effects through ablation studies, as well as the relationship between the number of groups and the semantic integrity in each group. Comparable performance to the state-of-the-art methods and low computational cost make it possible to be widely employed in FGIC applications. 

\bibliographystyle{IEEEtran}
\bibliography{mybib}

\section*{Appendix A: Data statistics}
We validate the effectiveness of our approach on 4 fine-grained benchmark datasets. Detailed instructions of these datasets can be found on their homepages, respectively. The data statistics are presented in the following Table.
\begin{table}[h]
    \centering
    \caption{The statistics of fine-grained datasets in this letter}
    \begin{threeparttable}
        \begin{tabular}{lccc}
            \toprule
            Datasets & $\#$category & $\#$training & $\#$testing \\
            \midrule
            CUB-Birds~\cite{cubbirds11caltech} & $200$ & $5,994$ & $5,794$ \\
            Stanford Cars~\cite{stcars13feifei} & 196 & 8,144 &8,041 \\
            Stanford Dogs~\cite{stdogs11feifei} & 120 & 12,000 &8,580\\
            FGVC-Aircraft~\cite{vggaircraft13Vedaldi} & 100 & 6,667 &3,333\\
            \bottomrule
        \end{tabular}
    \end{threeparttable}
    \label{tab:datasets}
\end{table}

\section*{Appendix B: Structures of backbone networks and Training Details}
\textbf{VGG16\cite{vggnet15zisserman}.} Due the the huge parameters of VGG16, we correspondingly modified the structure of VGG16 to make it suitable for our validation. To this end, we remove all the fully connected layers and employ the global average pooling on the last convolutional feature maps, then directly feed the pooled features into the output layer. This modification reduces about $\mathbf{89\%}$ parameters of the vanilla VGG16 model. Our approach introduces about $\mathbf{0.7\%}$ additional parameters to this modified VGG16 on the Birds dataset.
\\

\textbf{ResNeXt-50\cite{resnext17kaiming}.} We use the default setting of ResNeXt-50, which has 32 paths in all blocks. The input and output channels in every path are 4-dimensional. This setting ensures the similar capacity of ResNeXt-50 and ResNet-50. Thus, our approach also introduces about $\mathbf{1.7\%}$ additional parameters to ResNeXt-50 on the Birds dataset.
\\

\textbf{Training details.} We did not fine-tune hyper-parameters (including the number of groups, learning rate, regularization coefficients) for VGG16 and ResNeXt-50, but used the same hyper-parameters as ResNet-50. Thus, better results can be achieved if fine-tuning these hyper-parameters on target datasets. Both backbones were initialized by using pretrained weights on ImageNet~\cite{imagenet@feifei}. All other training criteria are the same as ResNet-50.
\end{document}